\documentclass[10pt,twocolumn,letterpaper]{article}

\usepackage[pagenumbers]{cvpr} 

\usepackage{graphicx}
\usepackage{amsmath}
\usepackage{amssymb}
\usepackage{booktabs}

\usepackage[pagebackref,breaklinks,colorlinks]{hyperref}

\usepackage[capitalize]{cleveref}
\crefname{section}{Sec.}{Secs.}
\Crefname{section}{Section}{Sections}
\Crefname{table}{Table}{Tables}
\crefname{table}{Tab.}{Tabs.}

\usepackage{multirow}

\usepackage[accsupp]{axessibility}  

\def\cvprPaperID{1559} 
\def\confName{CVPR}
\def\confYear{2022}

\begin{document}

\title{PhotoScene: Photorealistic Material and Lighting Transfer for Indoor Scenes}

\author{{Yu-Ying Yeh}\textsuperscript{1} \quad {Zhengqin Li}\textsuperscript{1} \quad {Yannick Hold-Geoffroy}\textsuperscript{2} \quad {Rui Zhu}\textsuperscript{1} \quad {Zexiang Xu}\textsuperscript{2}  \\ 
\quad {Miloš Hašan}\textsuperscript{2} \quad {Kalyan Sunkavalli}\textsuperscript{2} \quad {Manmohan Chandraker}\textsuperscript{1}\\[2mm]
{$^{1}$University of California, San Diego }\quad {$^{2}$Adobe Research}
}

\maketitle

\begin{abstract}
\vspace{-0.1cm}
Most indoor 3D scene reconstruction methods focus on recovering 3D geometry and scene layout. In this work, we go beyond this to propose PhotoScene\footnote{Code: \url{https://github.com/ViLab-UCSD/photoscene}}, a framework that takes input image(s) of a scene along with approximately aligned CAD geometry (either reconstructed automatically or manually specified) and builds a photorealistic digital twin with high-quality materials and similar lighting. We model scene materials using procedural material graphs; such graphs represent photorealistic and resolution-independent materials. We optimize the parameters of these graphs and their texture scale and rotation, as well as the scene lighting to best match the input image via a differentiable rendering layer. We evaluate our technique on objects and layout reconstructions from ScanNet, SUN RGB-D and stock photographs, and demonstrate that our method reconstructs high-quality, fully relightable 3D scenes that can be re-rendered under arbitrary viewpoints, zooms and lighting.

\vspace{-0.3cm}

\end{abstract}

\section{Introduction}

A core need in 3D content creation is to recreate indoor scenes from photographs with a high degree of photorealism. Such photorealistic ``digital twins'' can be used in a variety of applications including augmented reality, photographic editing and simulations for training in synthetic yet realistic environments. In recent years, commodity RGBD sensors have become common and remarkable progress has been made in reconstructing 3D scene geometry from both single \cite{fernandez2018layouts,sun2019horizonnet} and multiple photographs \cite{yao2019recurrent}, as well as in aligning 3D models to images to build CAD-like scene reconstructions \cite{izadinia2017im2cad,avetisyan2020scenecad,huang2018holistic,nie2020total3dunderstanding}. But photorealistic applications require going beyond the above geometry acquisition to capture material and lighting too — to not only recreate appearances accurately but also visualize and edit them at arbitrary resolutions, under novel views and illumination. 

Prior works assign material to geometry under the simplifying assumptions of homogeneous material \cite{izadinia2017im2cad} or single objects \cite{photoshape2018}. In contrast, we deal with the challenge of ascribing spatially-varying material to an indoor scene while reasoning about its complex and global interactions with arbitrary unknown illumination. One approach to our problem would be to rely on state-of-the-art inverse rendering methods \cite{li2020inverse, li2020openrooms} to reconstruct per-pixel material properties and lighting. However, these methods are limited to the viewpoint and resolution of the input photograph, and do not assign materials to regions that are not visible (either outside the field of view or occluded by other objects). Instead, we posit that learned scene priors from inverse rendering are a good initialization, whereafter a judicious combination of expressive material priors and physically-based differentiable rendering can solve the extremely ill-posed optimization of spatially-varying material and lighting. 

\begin{figure}[t]
    \begin{center}
    \includegraphics[width=0.48\textwidth]{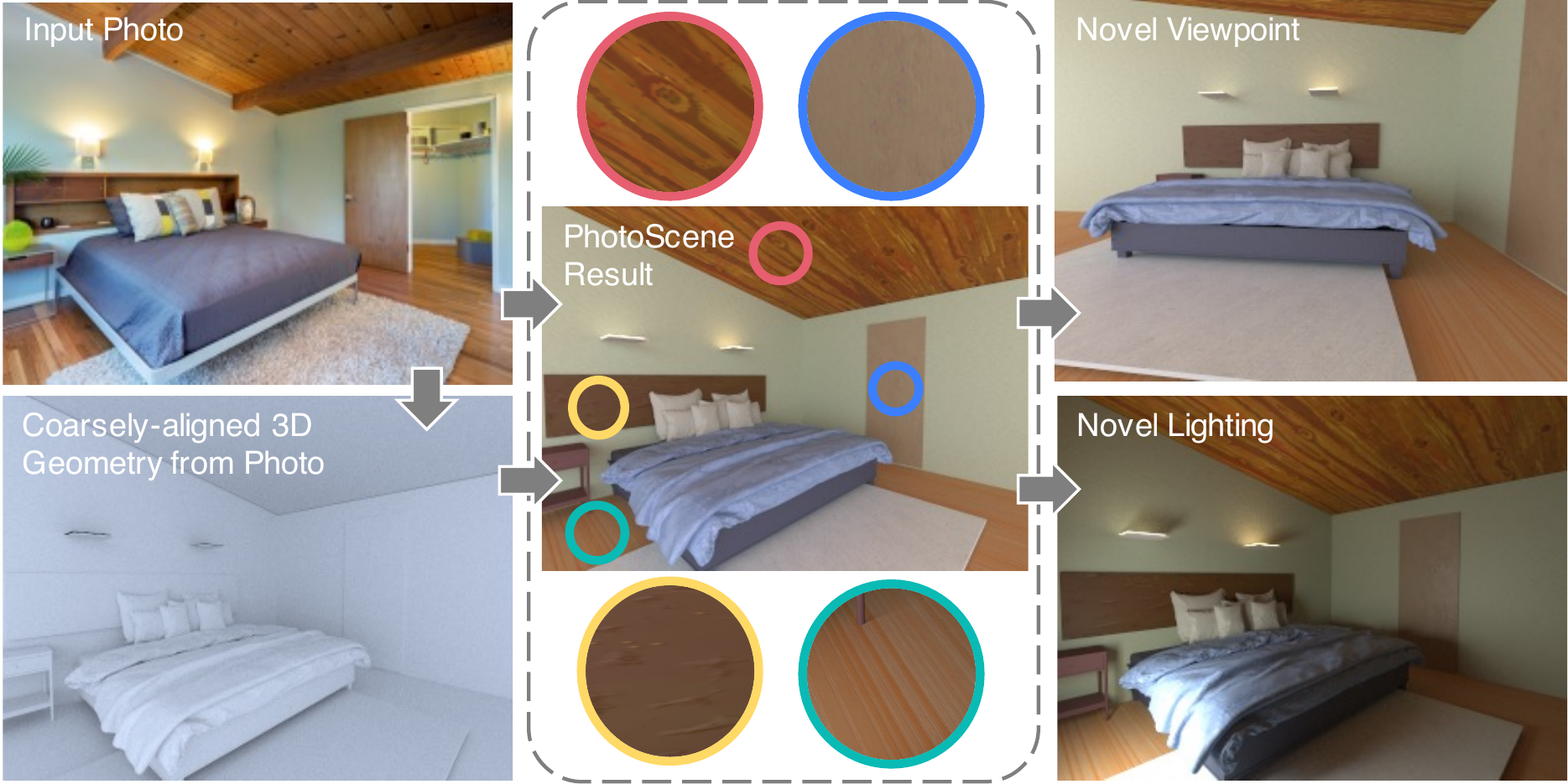}
	\end{center}
	\vspace{-0.4cm}
	\caption{Given an input photo and a coarsely aligned 3D scene model, 
	PhotoScene automatically infers high-quality spatially-varying procedural materials and scene illumination to closely match scene  appearance. The reconstructed materials are resolution-independent (see zoom insets) and ascribed to the full 3D geometry, to create a high-quality photorealistic digital twin that can be rendered under novel views and lighting. 
	}
	\vspace{-0.5cm}
	\label{fig:teaser}
\end{figure}

In this paper, we use procedural node graphs as compact yet expressive priors for scene material properties. Such graphs are heavily used in the content and design industry to represent \emph{high-quality, resolution-independent} materials with a compact set of optimizable parameters \cite{adobestock,substance}. This offers a significant advantage: if the parameters of a procedural graph can be estimated from just the observed parts of the scene in an image, we can use the full graph to ascribe materials to the entire scene. Prior work of Shi et al.~\cite{Shi2020:ToG} estimates procedural materials, but is restricted to fully observed flat material samples imaged under known flash illumination. In contrast, we demonstrate that such procedural materials can be estimated from partial observations of indoor scenes under arbitrary, unknown illumination.

We assume as input a coarse 3D model of the scene with possibly imperfect alignment to the image, obtained through 3D reconstruction methods \cite{avetisyan2020scenecad,huang2018holistic,nie2020total3dunderstanding}, or manually assembled by an artist. We segment the image into distinct material regions, identify an appropriate procedural graph (from a library) for each region, then use the 3D scene geometries and their corresponding texture UV parameterizations to ``unwarp'' these pixels into (usually incomplete) 2D textures. This establishes a fully differentiable pipeline from the parameters of the procedural material via a physically-based rendering layer to an image of the scene, allowing us to backpropagate the rendering error to optimize the material parameters. In addition, we also estimate rotation and scale of the UV parameterization and optimize the parameters of the globally-consistent scene illumination to best match the input photograph.

As shown in Fig.~\ref{fig:teaser} our method can infer spatially-varying materials and lighting even from a single image. Transferring these materials to the input geometry produces a fully relightable 3D scene that can then be rendered under novel viewpoint or lighting. Since procedural materials are resolution-invariant and tileable, we can render closeup views that reveal fine material details, without having observed these in the input photograph. This goes significantly beyond the capabilities of current scene-level inverse rendering methods and allows for the creation of high-quality, photorealistic replicas of complex indoor scenes.

\section{Related Works}

\noindent{\bf Material acquisition and recognition}
High-quality materials have been estimated in many prior works, using both single \cite{aittala2016,deschaintre2018single,li2018materials,henzler2021generative} or multiple \cite{deschaintre2019,gao2019} input images. Most of the above methods estimate materials for planar samples, as opposed our inputs that are unconstrained images of complex indoor scenes. While material recognition methods have been proposed to classify image regions into material categories \cite{bell15minc}, they do not yield parametric materials that could be used for relighting and view synthesis. In recent years, several methods have been proposed to use procedural graphs as materials priors and estimate their parameters to  match the appearance of captured images \cite{guo2019,hu2019,Shi2020:ToG}. In particular, we use MATch\cite{Shi2020:ToG}, a differentiable procedural material model based on Substance node graphs, to constrain our materials to the SVBRDF manifold. However, the above methods only consider flat material samples captured under known flash illumination, while our goal is significantly more challenging -- our inputs are photos of indoor scenes with complex geometry, lit by unknown spatially-varying illumination, with scene layout and occlusions leading to incomplete textures with arbitrary scales and rotation.

\vspace{0.1cm}
\noindent{\bf Inverse rendering of indoor scenes} 
Our problem may be seen as an instance of inverse rendering \cite{Marschner1998InverseRF,patow2003}, but we must estimate materials that are amenable for rendering under novel views and lighting, as well as editability. Several approaches have been proposed for inverse rendering for objects \cite{liu2017material,sengupta2018sfsnet,shu2017neural,tewari2017mofa,li2018learning,sang2020relighting}, but our focus is on indoor scenes, which have been considered in both early \cite{barron2013intrinsic,karsch2011rendering,karsch2014automatic} and recent \cite{li2020inverse,sengupta2019nir} works. A convolutional neural network with a differentiable rendering layer estimates depths, per-pixel SVBRDF and spatially-varying lighting in \cite{li2020inverse} using a single input image. We use their material and lighting outputs as our initialization and to aid our differentiable rendering losses in image space, but go further to assign procedural materials that can be used for rendering novel views and estimate lighting that is consistent across views. Recent work has applied differentiable rendering \cite{Li:2018:DMC} to recover spatially-varying reflectance and lighting given photos and 3D geometry \cite{azinovic2019inverse,nimier2021material}. However, these methods estimate per-vertex BRDFs and require high-quality geometry as input, while our procedural models regularize scene materials, allowing us to infer them from only coarsely aligned 3D models and to re-render novel viewpoints with material detail that was not observed in the input photos.

\begin{figure*}
\begin{center}
\includegraphics[width=1.\linewidth]{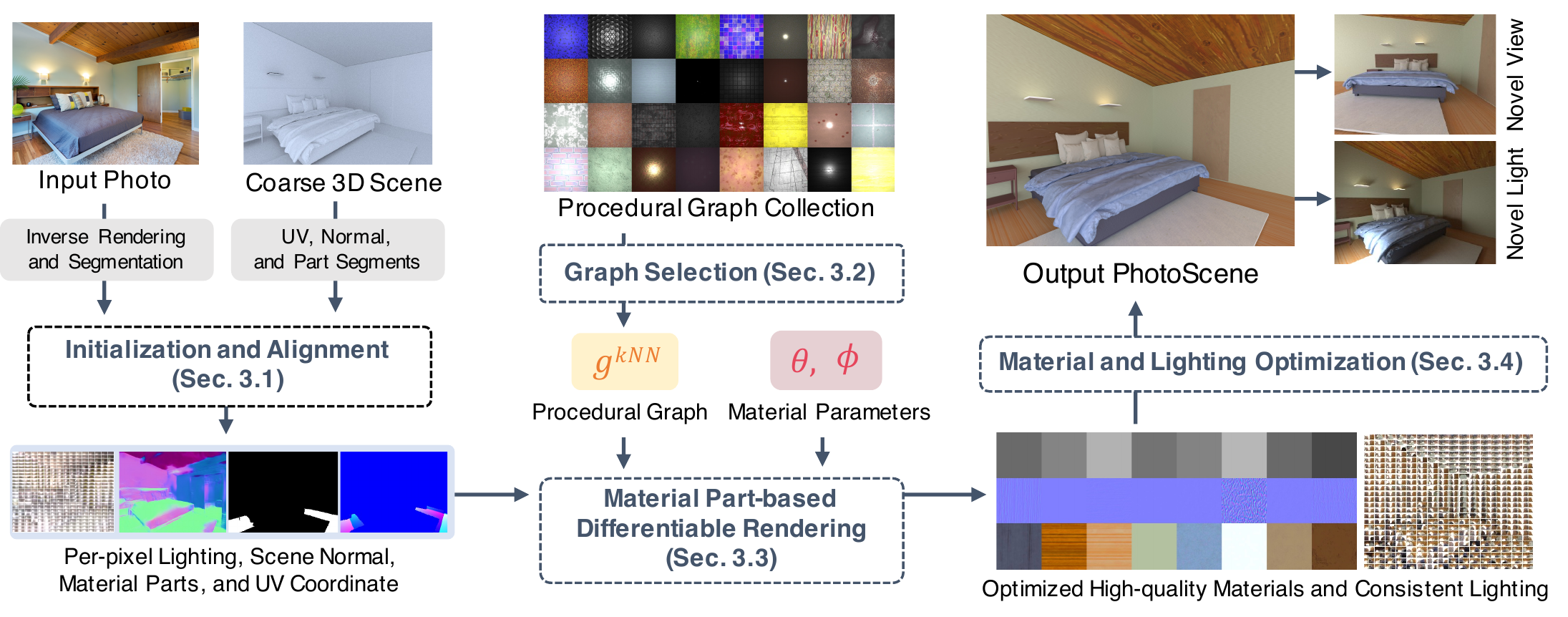}
\end{center}
\vspace{-0.7cm}
\caption{The PhotoScene framework. From the input photo(s) and 3D scene model, we estimate scene normals and lighting via an inverse rendering method, and compute material parts and align them to the model UVs and part segments. We model scene materials with procedural graphs. For each material part, we identify an appropriate graph from a collection and use a differentiable rendering module to optimize for the graph and UV transformation parameters. We also refine the initial lighting. Assigning the optimized materials and lighting to the input 3D model gives us our output PhotoScene---a renderable 3D scene that matches the appearance of the input photos.
}

\vspace{-0.3cm}
\label{fig:method}
\end{figure*}

\noindent{\bf Material transfer from photographs}
LIME~\cite{meka2018lime} clones the homogeneous material from a single color image.
Material Memex \cite{jain2012material} and Unsupervised Texture Transfer \cite{wang-utt} exploit correlations between part geometries and materials, or across patches for material transfer to objects. 
PhotoShape \cite{photoshape2018} assigns photorealistic materials to 3D objects through material classifiers trained on a synthetic dataset. In contrast, we seek material transfer in indoor scenes, which is significantly harder since image appearances entwine material properties with complex light transport. The material suggestion system of \cite{magicdecorator} textures synthetic indoor scenes with high-quality materials, but based on a set of pre-defined rules for local material and global aesthetics, whereas we must solve challenging inverse problems in a differentiable framework to match the appearances of real images.

\noindent{\bf Indoor scene 3D reconstruction}
Many works reconstruct indoor 3D scene geometry (objects and room layout) from single images \cite{izadinia2017im2cad,huang2018holistic,nie2020total3dunderstanding} or RGBD scans \cite{avetisyan2020scenecad}, but either do not address material and lighting estimation, or use heuristics like median color to assign diffuse textures \cite{izadinia2017im2cad}. Our work focuses on reconstructing high-quality material and lighting from input photos and is complementary to geometric reconstruction methods; indeed, we can leverage these methods to build our input coarse 3D model. Like us, Plan2Scene \cite{Vidanapathirana2021Plan2Scene} also aims to reconstruct textured 3D scenes, but is limited to diffuse textures and constrained by the quality of the texture synthesis model. In contrast, by optimizing a procedural material model, we are able to recover high-quality non-Lambertian materials; we also optimize for illumination and hence better match the input image appearance.

\section{Proposed Method}
\vspace{-0.2cm}

Our method starts from an input image (or multiple images) of an indoor scene and a roughly matching scene reconstruction (automatic or manual). Our goal is to obtain high-resolution tileable material textures for each object, as well as a globally consistent lighting for the scene. 

The method consists of four high-level stages, as shown in Fig.~\ref{fig:method}. We compute an initial estimate of scene normals and lighting. We also find material parts, and align them between the input and rendered image, so that each material part can be optimized separately. Next, we choose a \emph{material prior} for each material part, in the form of a procedural node graph that produces the material's textures (albedo, normal and roughness) given a small set of parameters. Finally, we optimize the parameters of all materials as well as the lighting in the scene. 
Below we describe this in detail.

\subsection{Initialization and Alignment}
\label{sec:init}
\vspace{-0.2cm}

In the initialization step, we obtain estimates of normals and lighting from the input image(s) that guide the subsequent optimization. 
Next, since the synthetic scene is composed of elements that are not perfectly aligned with the input photograph(s), we warp the pixels rendered from the geometry to best fit the scene structure in the input photograph \textit{per material part}. If there are multiple input images per scene available, we perform consensus-aware view selection (see Sec.~\ref{supp:consensus_select} for details).

\vspace{0.1cm}
\noindent\textbf{Pixel-level normals and lighting initialization.}
We use the pretrained inverse rendering network (InvRenderNet) from Li et al. \cite{li2020inverse} to obtain spatially-varying incoming lighting estimates  $\mathcal{L}^\text{inv}$ and per-pixel normals $\mathcal{N}^\text{inv}$ to guide the material optimization in pixel space (Sec.~\ref{sec:diffrender}).
We do not use the estimated albedo and roughness from InvRenderNet, except as baselines for comparison, as shown in Fig.~\ref{fig:compare}.

\vspace{0.1cm}
\noindent\textbf{Material part mask and mapping.} 
For each material part, our method requires a mask $M_\text{photo}$ to indicate the region of interest in the input image, and another mask $M_\text{geo}$ to indicate the same in the synthetic image. The latter mask is trivially available by rendering the synthetic geometry. To obtain $M_\text{photo}$ automatically, we make use of predictions from MaskFormer~\cite{cheng2021per} as proposals, and find the mapping by computing maximum intersection over union (IoU) with respect to $M_\text{geo}$ of all material parts. Semantic and instance labels (when available) can be used to reduce the proposals. To obtain more robust results, manually segmented masks can be taken as $M_\text{photo}$ instead. More details can be found in Sec.~\ref{supp:mask}.

\vspace{0.1cm}
\noindent\textbf{Geometry/photo alignment and warping.} \label{sec:alignment}
To handle misalignment between material parts in the input image and the geometry, we compute an affine pixel warp. We obtain pixel locations $\mathbf{x}_s^*$ to sample values from $M_\text{geo}$ of the geometry to warp to the material mask $M_\text{photo}$ at pixel $\mathbf{x}_t$ as
\begin{equation}
    \mathbf{x}_s^* = \tfrac{\mathbf{x}_t - \mathbf{c}_{p}}{\mathbf{l}_{p}} \cdot \mathbf{l}_{g} + \mathbf{c}_{g} \; ,
\label{eq:ptopt}
\end{equation}
where $\{ \mathbf{c}_{g}, \mathbf{l}_{g} \}$ and $\{ \mathbf{c}_{p}, \mathbf{l}_{p} \}$ are the centers and sizes of bounding boxes around the masks. As the affine warp is imperfect, some pixels will not have a correspondence and will be dropped. Please see Sec.~\ref{supp:alignment} for details.

At the end of this initialization and alignment step, we obtain per-material part UVs and corresponding masks in the input and synthetic images, as well as an initial estimate of scene lighting and normals. 

\subsection{Material Prior: Procedural Node Graphs}
\label{sec:matprior}
\vspace{-0.1cm}

\begin{figure}[t]
\centering
\vspace{-0.1cm}
\includegraphics[width=0.9\linewidth]{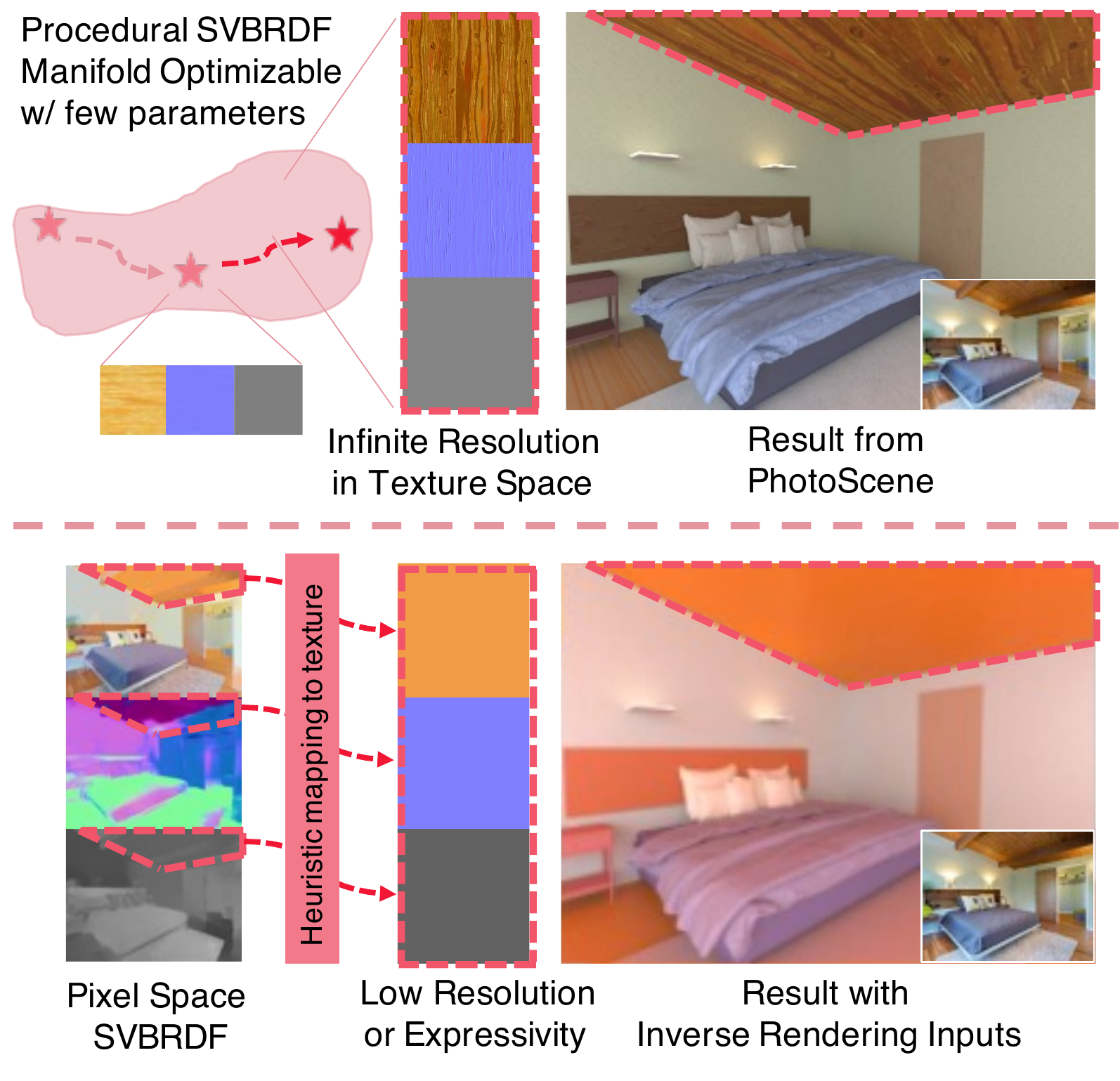}
    \vspace{-0.3cm}
  \caption{An expressive material prior optimizable with a few parameters allows recreating photorealistic appearances, in contrast to heuristics that may rely solely on pixel-space inverse rendering.}
   \vspace{-0.3cm}
\label{fig:procedural}
\end{figure}
Modeling spatially-varying materials as 2D textures is difficult due to the ill-posed nature of the problem: the textures may not be fully observed in the input image, and they are lit by uncontrolled illumination. Therefore, a key step of our method is to constrain materials to lie on a valid SVBRDF manifold, by specifying a \emph{material prior} that is expressive, yet determined by a small number of parameters. This material prior must be differentiable to allow parameter optimization via backpropagation. 
This is in contrast to a bottom-up approach that might rely on pixel space outputs from inverse rendering (see Fig.~\ref{fig:procedural} and~\ref{fig:compare}).

We use MATch \cite{Shi2020:ToG}, a material prior based on differentiable procedural node graphs. Their implementation provides 88 differentiable procedural graphs that model high quality spatially-varying materials, each with a unique set of parameters. For our purposes, these graphs are simply differentiable functions from a parameter vector $\theta$ to albedo, normal and roughness textures $A_{uv}$, $N_{uv}$, $R_{uv}$. We add an additional offset parameter for the albedo output from the graphs to more easily control the dominant albedo colors.  We select 71 graphs that are representative of indoor scenes as our graph collection $\{g^1, ..., g^{71}\}$. We augment this set with a homogeneous material, used for untextured parts.

\vspace{0.1cm}
\noindent\textbf{Material graph selection.} For each material part in the image, we need to choose an appropriate procedural node graph from the library. We address this by nearest neighbor search using a VGG feature distance. Specifically, we sample 10 materials from each of the 71 graphs with random parameters, resulting in 710 exemplar material maps $\{(A|N|R)_{uv}^1, \cdots,  (A|N|R)_{uv}^{710}\}$. We render the part using each exemplar with our differentiable renderer (Sec.~\ref{sec:diffrender}), resulting in 710 render-graph pairs $\{(\mathcal{\Tilde{I}}_{rend}^1, g^1), (\mathcal{\Tilde{I}}_{rend}^2, g^1), ... (\mathcal{\Tilde{I}}_{rend}^{710}, g^{71})\}$, then select the $k$ ($=21$) most similar renderings to the input using a masked VGG distance (Eq.~\ref{eq:vgg}), which vote for their corresponding graphs and we pick the graph $g^{kNN}$ with the most votes. 

\begin{figure}[!t]
\centering
\includegraphics[width=0.98\columnwidth]{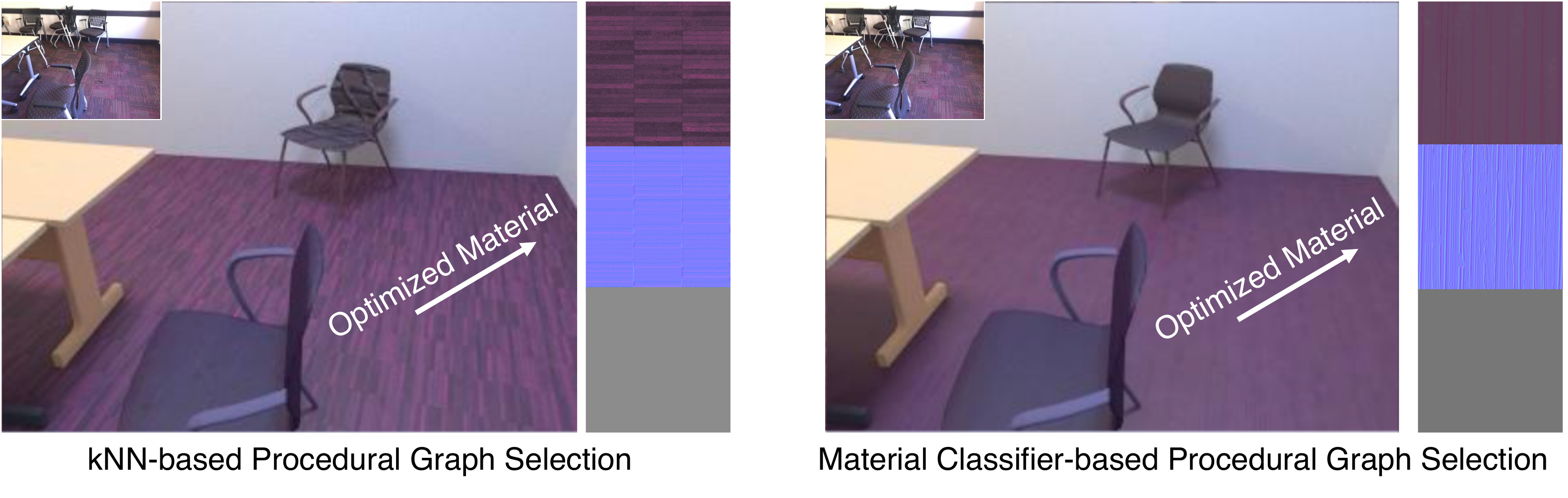}
\vspace{-0.3cm}
\caption{\small Graph selection with kNN versus material classifier.
} 
\label{fig:knn_vs_matcls}

\end{figure}

We also experiment with predicting material super classes (e.g. wood, plastic, etc.) using a pretrained classifier and then selecting a graph from the class, but find the kNN search less susceptible to errors (Fig.~\ref{fig:knn_vs_matcls}). 
For small parts where it is difficult to observe spatial variations, we use homogeneous materials.

\subsection{Material Part Differentiable Rendering}
\label{sec:diffrender}

\vspace{-0.2cm}
\begin{figure}[t]
\centering
\includegraphics[width=\linewidth]{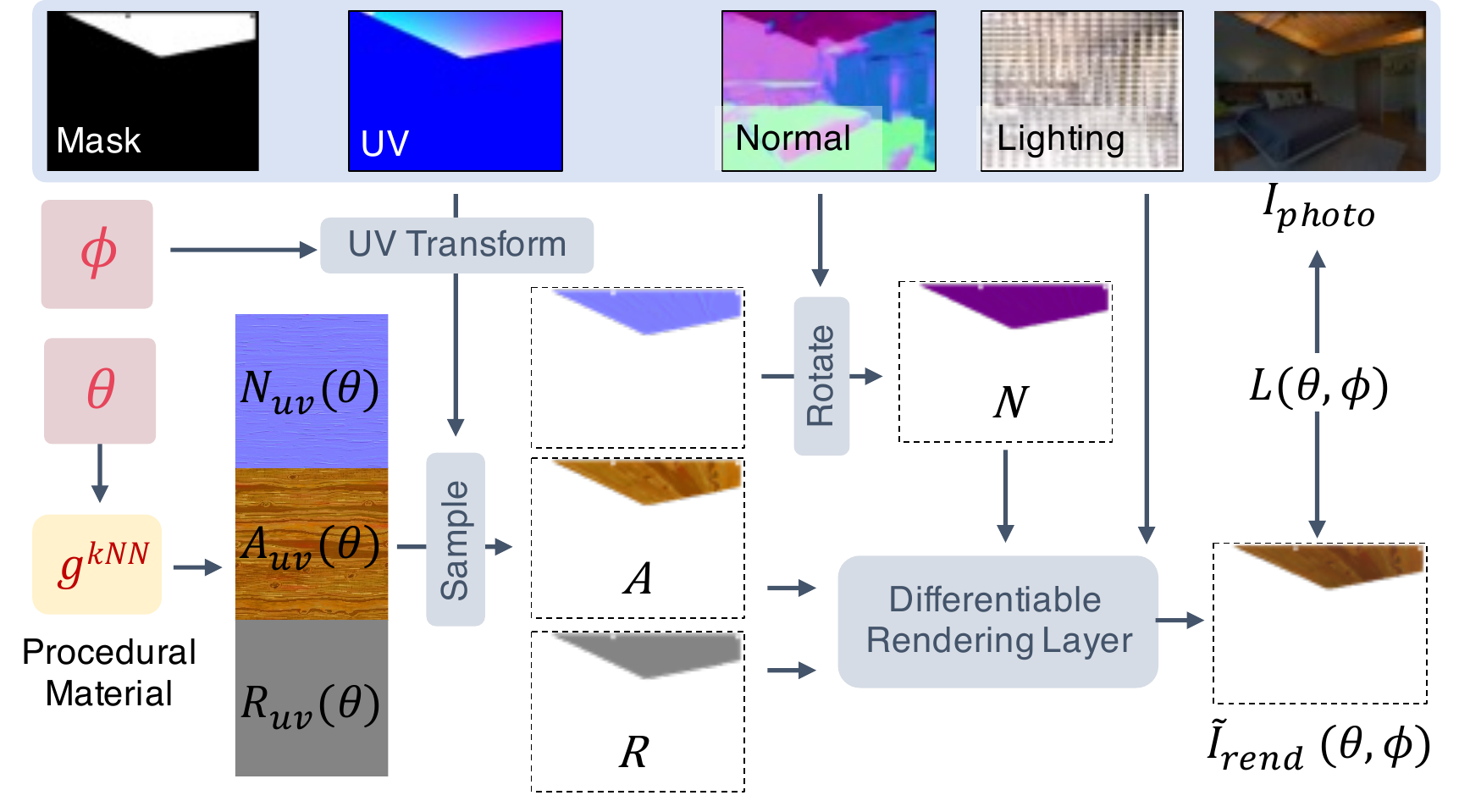}
   \vspace{-0.6cm}
   \caption{Given a material part mask, UV, scene normals and lighting (top), we construct a fully differentiable pipeline from material graph ($\theta$) and UV transformation ($\phi$) parameters via a texture-to-image mapping and differentiable rendering layer to a rendered image. We optimize for these parameters by comparing this rendering to the input photo.}
   \vspace{-0.4cm}
\label{fig:renderLayer}
\end{figure}

Our framework leverages a \emph{material part differentiable rendering module} as a way to predict the appearance of a part given its texture-space material maps $(A|N|R)_{uv}$, pixel-space geometry $\mathcal{N}^\text{inv}$, and local lighting $\mathcal{L}^\text{inv}$.
The differentiable rendering module can be used for optimization through back-propagation. 
We use it to optimize for material parameters in Sec.~\ref{sec:optim}.

During rendering, we use a spatially-varying grid of incoming light environment maps \cite{li2020inverse} as our lighting representation, which allows the operation of the rendering module to remain local; no additional rays need to be traced by the rendering process, which is crucial for efficiency.

Our differentiable rendering module is shown in Fig.~\ref{fig:renderLayer}.
Our material model is a physically-based microfacet BRDF~\cite{karis2013real}. The rendering module takes per-pixel texture (UV) coordinates, to sample material textures $A_{uv}$, $N_{uv}$, $R_{uv}$ generated from the material prior using parameters $\theta$ as $g^{kNN}(\theta)$. The normals need to be rotated into the local shading frame of a given point on the material part. The rendering module then uses per-pixel material parameters $A$, $N$, $R$ and spatially-varying local incoming lighting $L$ to render the image pixels $\Tilde{I}_\text{rend}$:
\begin{eqnarray}
    A, R =& \mathbf{Sample}_{UV}(A_{uv}(\theta), R_{uv}(\theta) ), \\
    N =& \mathbf{Rot} ( \mathbf{Sample}_{UV}(N_{uv}(\theta)) ), \\
    \mathcal{\Tilde{I}}_\text{rend} =& \mathbf{RenderLayer}(A, N, R, L).
\end{eqnarray}
As the originally assigned UV coordinates might not have the optimal scale and orientation to apply the corresponding material, we apply texture transformation parameters $\phi$ (rotation, scale, and translation) to map original coordinates ${UV}^0$ to more appropriate ones ${UV}$.
\begin{eqnarray}
    UV =& \mathbf{UVTransform}({UV}^0, \phi).
\end{eqnarray}
Fig.~\ref{fig:results_hq} provides visual examples of the importance of considering rotation and scale in our differentiable rendering.

\subsection{Material and Lighting Optimization}
\label{sec:optim}

Images conflate lighting, geometry, and materials into an intensity value.
To better disambiguate lighting from material, we adopt a two-step process. First, we use our initial spatially-varying lighting prediction from InvRenderNet to optimize the materials for each object. Next, we perform a globally consistent lighting optimization to refine our illumination. Finally, we optimize the materials once more, this time using our refined lighting. This procedure reduces the signal leakage between material and lighting.

\vspace{0.1cm}
\noindent\textbf{Material optimization.} 
There is no exact correspondence between the rendered and reference pixels. Thus, we compute the absolute difference $\mathcal{L}_{stat}$ of the statistics (mean $\mu$ and variance $\sigma^2$) of the masked pixels of the part of interest to optimize both material prior parameters $\theta$ and UV transformation parameters $\phi$:
\begin{align}
    \mathcal{L}_\text{mean} &= | \mu(\mathcal{I}_\text{photo} \cdot M_\text{aln}) - \mu(\Tilde{\mathcal{I}}_\text{rend} \cdot M_\text{aln} )|, \\
    \mathcal{L}_\text{var} &= | \sigma^2(\mathcal{I}_\text{photo} \cdot M_\text{aln} ) - \sigma^2(\Tilde{\mathcal{I}}_\text{rend} \cdot M_\text{aln} )|,  \\
    \mathcal{L}_\text{stat} &= \mathcal{L}_\text{mean} + \mathcal{L}_\text{var},
\end{align}
where $M_\text{aln}$ is the resulting aligned mask from the alignment step.

Using this statistics loss encourages matching color distributions but not spatially-varying patterns. 
To further match the patterns, we add a masked version of VGG loss $L_{vgg}$~\cite{Simonyan15}. Let $\Tilde{C}_{l}$ and $C_l$ be the normalized VGG feature maps of $\mathcal{\Tilde{I}}_\text{rend}$ and $I_\text{photo}$ extracted from layer $l$\footnote{The layers used here are relu1\_2, relu2\_2, relu3\_3, relu4\_3, relu5\_3.}, we apply mask $M_{aln}$ on the sum of upsampled L2 difference of normalized feature maps $\Tilde{C}_{l}$ and $C_l$ and compute the mask-weighted average among the pixels $x$:
\begin{align}
\label{eq:vgg}
    \mathcal{L}_\text{vgg} = \frac{1}{\sum\limits_{x} {M_\text{aln}}}\sum_x M_\text{aln} \left( \sum_{l}\text{\small Up}\left( \mathcal{\Tilde{C}}_{l}-\mathcal{C}_{l} \right)^2 \right).
\end{align}
We also try a masked style loss based on the Gram matrices of VGG features~\cite{gatys2016image}, but find that it does not provide significant improvement over $L_\text{stat}$ and $L_{vgg}$. Therefore, we use the following loss for material optimization:
\begin{align}
    \mathcal{L}_\text{total} &= \alpha \mathcal{L}_\text{stat} + \beta \mathcal{L}_\text{vgg}.
\end{align}
Rather than jointly optimizing for material and UV parameters, we find that convergence is more stable with alternately searching for UV parameters in a discretized space and optimizing for material graph parameters.
Spatially-varying roughness parameters are difficult to optimize in a single view due to limited observations of highlights, so we replace the roughness output from the graph with a single mean value during optimization (the final result can still use the full roughness textures).

\vspace{0.1cm}
\noindent\textbf{Globally consistent lighting optimization.}
\begin{figure}
\begin{center}
\includegraphics[width=0.99\linewidth]{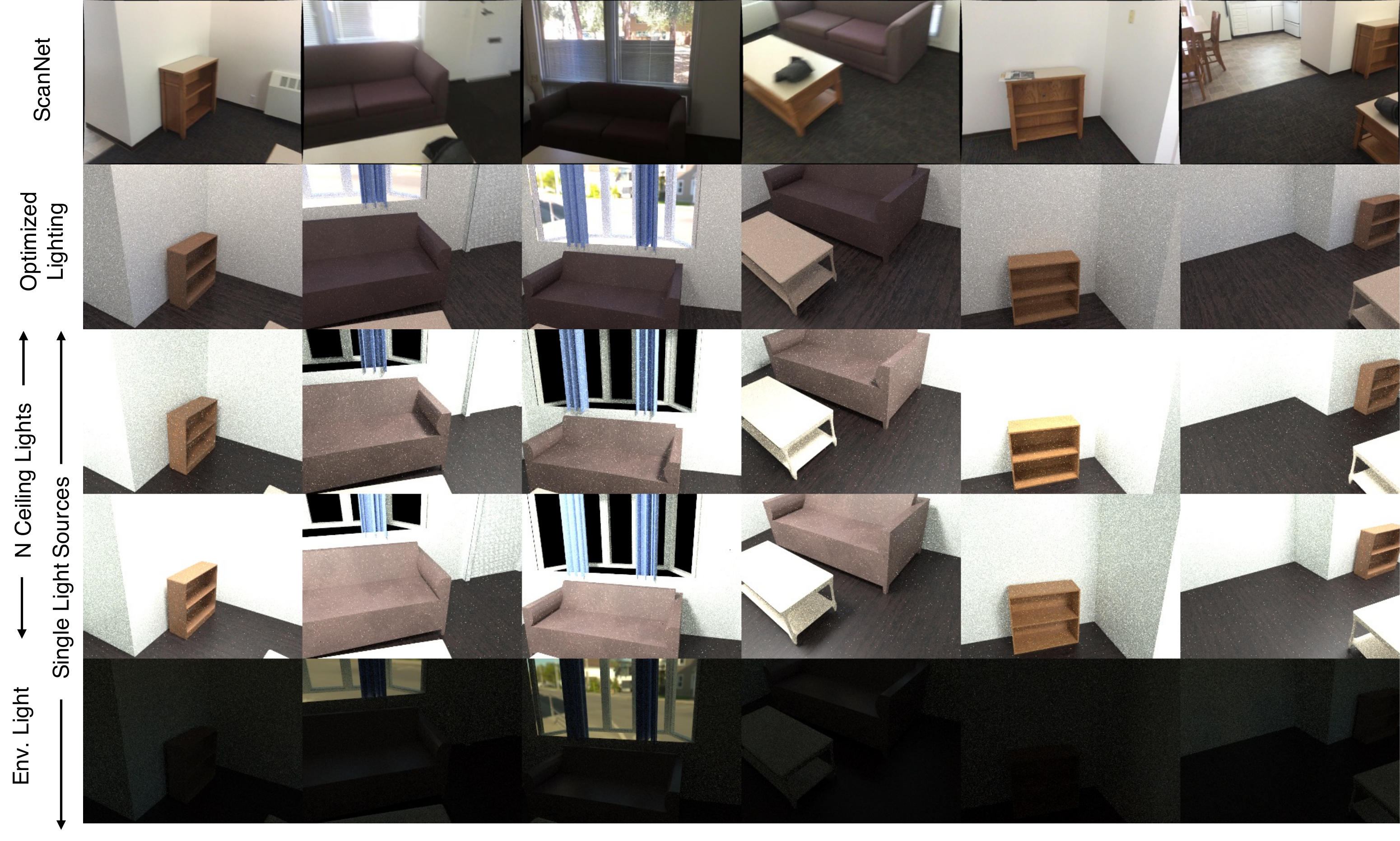}
\end{center}
\vspace{-0.5cm}
   \caption{Example of lighting optimization result using $N=2$ ceiling lights and one environment map lighting. The optimized lighting is close to original ScanNet images.}
\vspace{-0.3cm}
\label{fig:lighting}
\end{figure}
To estimate globally consistent lighting, we represent indoor lighting as $N$ area lights and one environment light which may be observed through the windows. 
We optimize for RGB intensities for each light source. For scenes without light source annotations, we uniformly place area lights on the ceiling every 3 meters of distance. 
With the materials we previously optimized, we render images with each single light source turned on. We compute the RGB intensities by comparing these renderings with the input view using least squares. Specifically, with $N$ area lights (including ceiling lights and lamps) in a room with $V$ input images, we solve for $3\times(N+1)$ RGB coefficients $x_r, x_g, x_b \in \mathbb{R}^{N+1}$ (the $+1$ refers to an environment light visible through windows). 
We use these coefficients to re-weight the intensities of each light source. Fig.~\ref{fig:lighting} demonstrates examples of rendering under selected views with each light source and the final combined optimized lighting. We additionally optimize for relative exposure values under different views, since they may vary over a video acquired using commodity cameras.

\begin{figure}
\begin{center}
\includegraphics[width=0.9\linewidth]{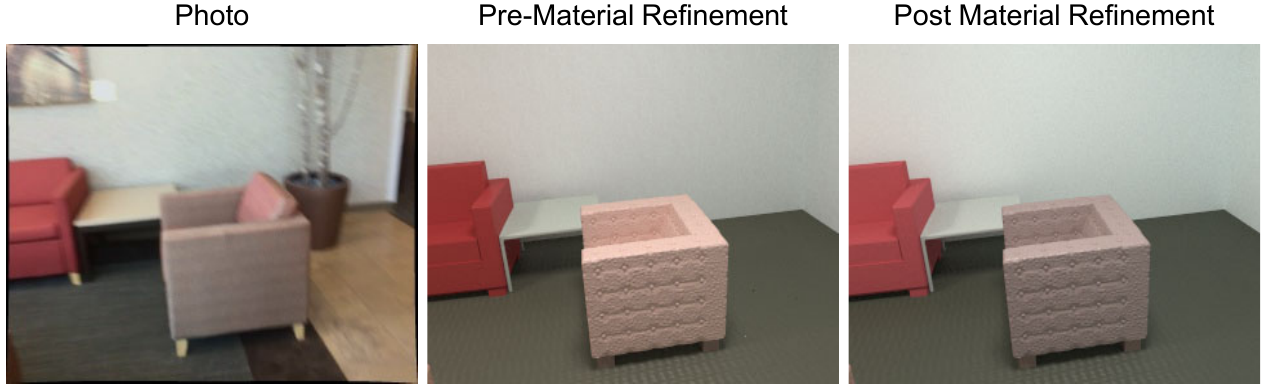}
\end{center}
\vspace{-0.5cm}
  \caption{Second-round material refinement successfully corrects the inaccurate reflectance values estimated in the first round. }
  \vspace{-0.4cm}
\label{fig:reoptim}
\end{figure}

\vspace{0.1cm}
\noindent\textbf{Material reoptimization.}
With the refined globally-consistent light sources, we re-optimize the materials to improve our results.
We render a new spatially-varying incoming lighting grid $L^\text{global}$ from the synthetic scene with the optimized light sources and use the same optimization loss as Sec.~\ref{sec:diffrender}. We only optimize for a homogeneous re-scaling of the albedo and roughness maps in this round, as the spatially-varying patterns are already correctly optimized by the first iteration. Fig.~\ref{fig:reoptim} demonstrates that this refinement step can rectify inaccurate material parameters caused by albedo-lighting ambiguities in the inverse rendering network.

\section{Experiments}

\subsection{Datasets}
\label{sec:dataset}

We demonstrate our method on photos and corresponding scene data from several sources and demonstrate its robustness on images and scene geometry of varying quality.

\noindent \textbf{ScanNet-to-OpenRooms.} 
We use geometry and 3D part segmentations from OpenRooms~\cite{li2020openrooms}, corresponding to \textit{multi-view} input images from ScanNet \cite{dai2017scannet} videos, with instance segmentation labels as mask proposals for $M_{photo}$.

\noindent \textbf{Photos-to-Manual.} 
For several high-quality real-world photos, we also manually construct matching scenes using Blender~\cite{blender} from a \textit{single view} and manually segment the material part masks for demonstration purposes.

\noindent \textbf{SUN-RGBD-to-Total3D.} 
Our method can also be used for fully automatic material and lighting transfer using a \textit{single-image} mesh reconstruction from Total3D~\cite{nie2020total3dunderstanding} applied to SUN-RGBD~\cite{song2015sun} inputs. The reconstruction in this case is coarser than CAD retrieval and with a single material per object. We use MaskFormer~\cite{cheng2021per} to obtain segmentation labels as mask proposals and use object classes for mapping. 

We assume all meshes have texture coordinates. If not, we use Blender's~\cite{blender} Smart UV feature to generate them.

\begin{figure}
\begin{center}
\includegraphics[width=0.99\linewidth]{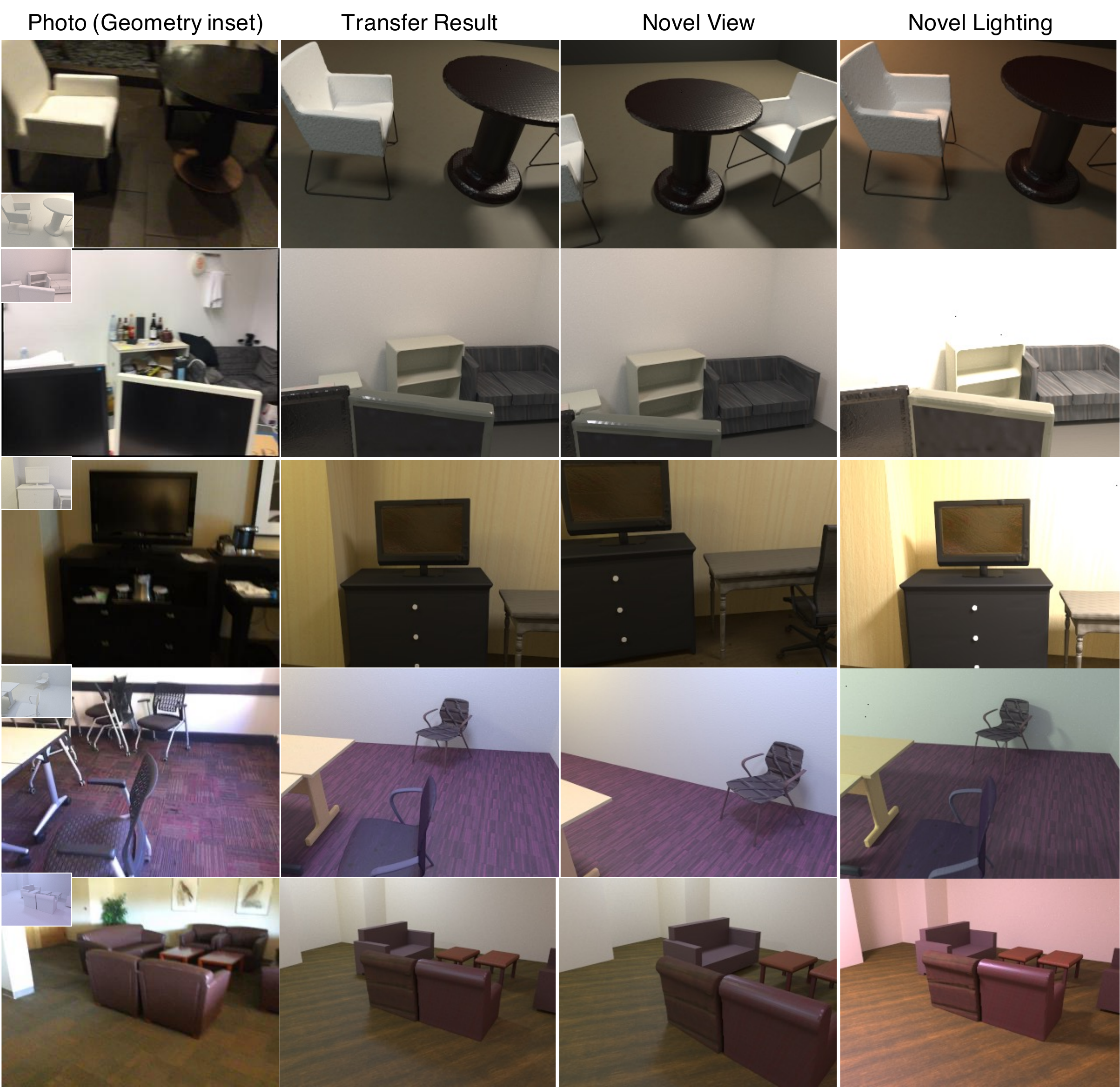}
\end{center}
\vspace{-0.5cm}
   \caption{Example of material transfer results for different scenes with \textit{ScanNet-to-OpenRooms}. }
\vspace{-0.3cm}
\label{fig:results_scannet}
\end{figure}

\begin{figure}
\begin{center}
\includegraphics[width=\linewidth]{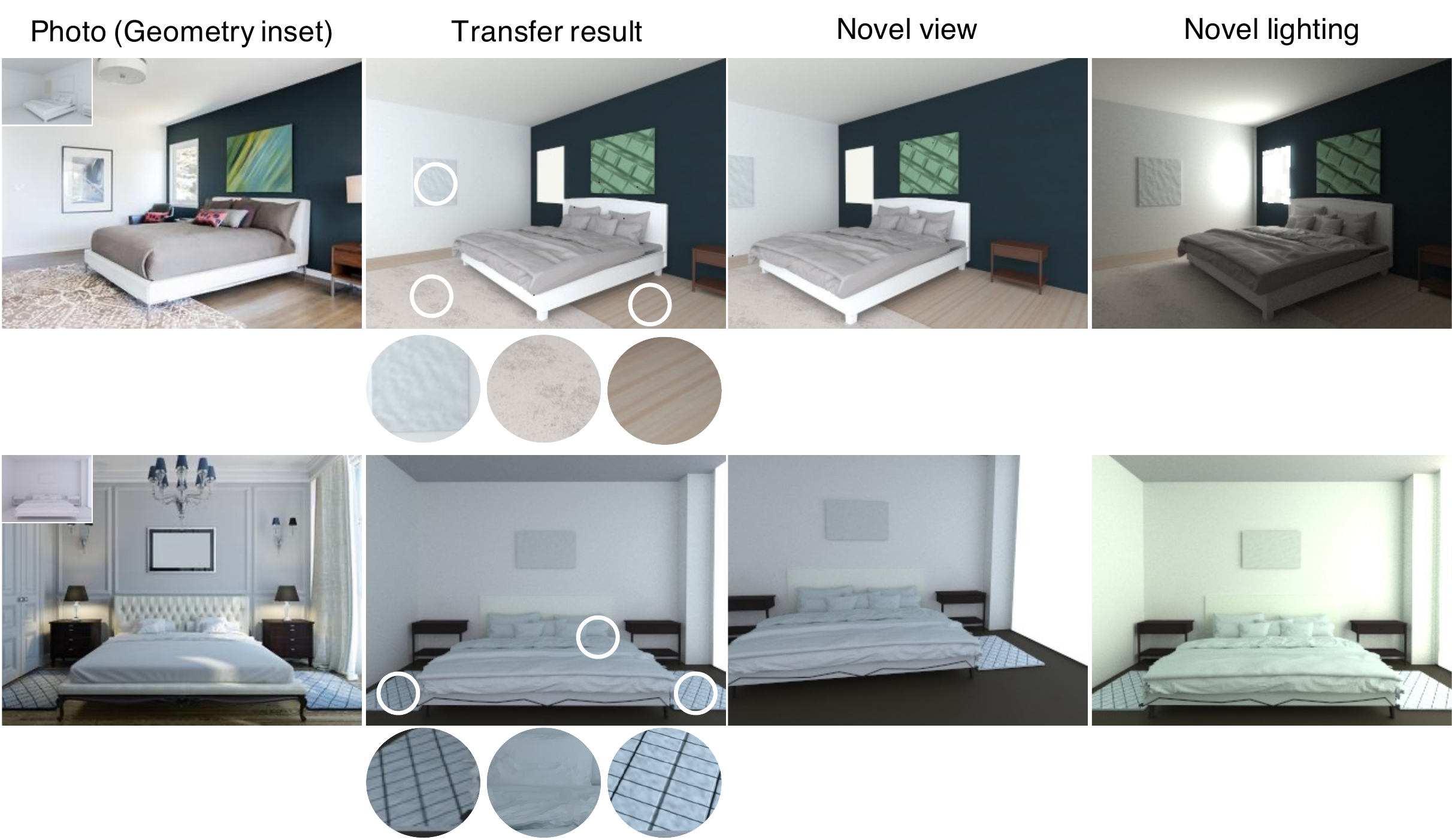}
\end{center}
\vspace{-0.5cm}
  \caption{Our material and lighting transfer results for two scenes in \textit{Photos-to-Manual} dataset. Note how our method is able to accurately reconstruct the appearance and orientation of the spatially-varying materials in these scenes.  }
\vspace{-0.3cm}
\label{fig:results_hq}
\end{figure}

\begin{figure}
\begin{center}
\includegraphics[width=0.99\linewidth]{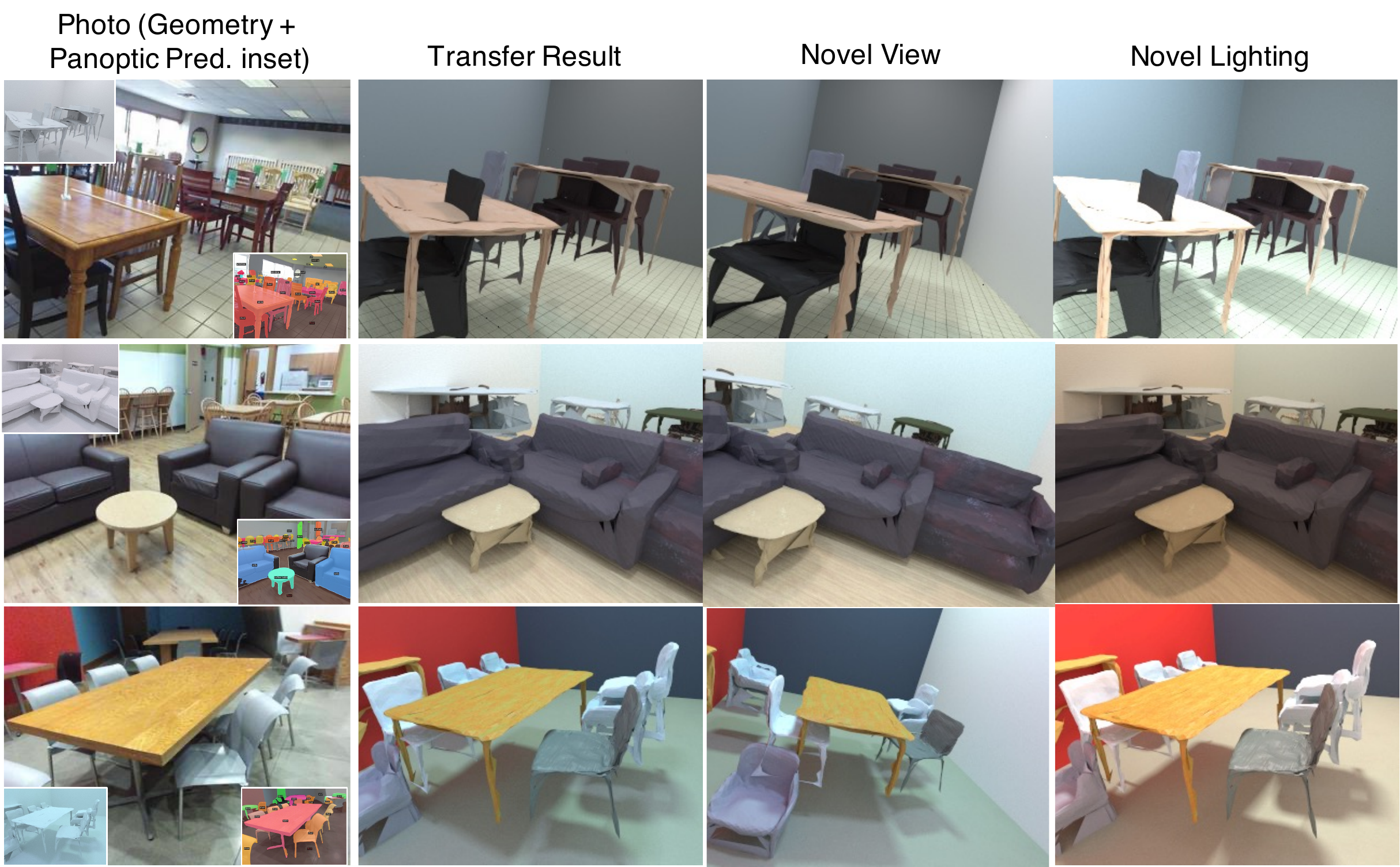}
\end{center}
\vspace{-0.5cm}
   \caption{Examples of material transfer results with \textit{SUN-RGBD-to-Total3D}. Note that these are fully automatic results by using Total3D to reconstruct 3D meshes and panoptic label predictions from MaskFormer. Our method is robust to imperfect geometry and panoptic prediction labels as shown in the inset. }
\vspace{-0.3cm}
\label{fig:results_total3d}
\end{figure}

\begin{figure}
\begin{center}
\includegraphics[width=0.99\linewidth]{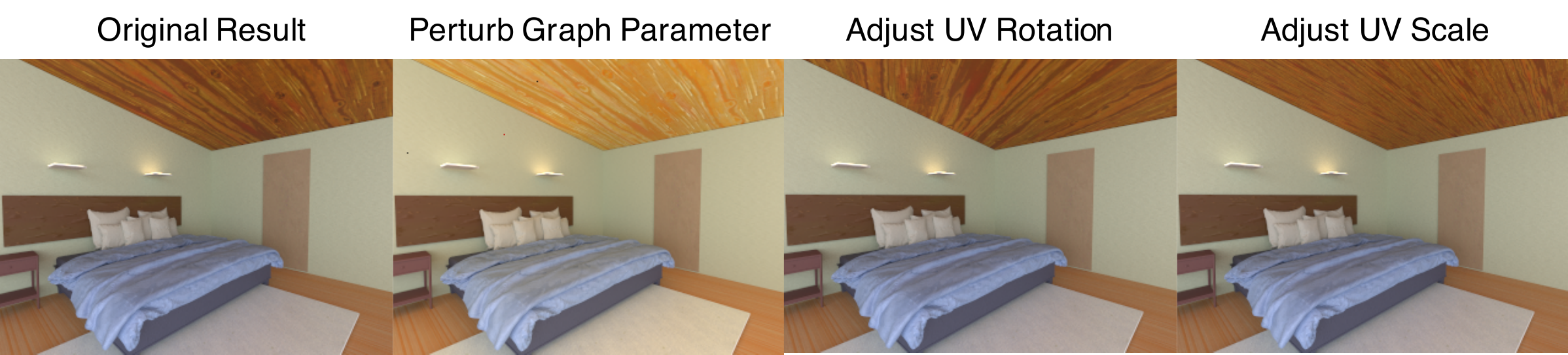}
\end{center}
\vspace{-0.5cm}
   \caption{Example of \textit{editable} variations from originally optimized procedural graph materials for ceiling. We can perturb graph parameters or adjust UV parameters from optimized results to generate various appearances.  }
\vspace{-0.3cm}
\label{fig:results_material_variations}
\end{figure}

\subsection{Material and Lighting Transfer}
\label{sec:result}
We demonstrate our material and lighting transfer results in Fig.~\ref{fig:results_scannet},~\ref{fig:results_hq} and~\ref{fig:results_total3d}, where our material prior allows interesting relighting effects such as specular highlights, shadows and global illumination under novel views and lighting.

\vspace{0.1cm}
\noindent\textbf{Multiview inputs with CAD geometry.}
We demonstrate the multi-view setup in Fig.~\ref{fig:results_scannet} with \textit{ScanNet-to-OpenRooms} dataset. We select an optimal view for every single material to get the best observation as well as to estimate the global lighting for the entire room. Even though the CAD models are neither perfectly aligned nor perfect replicas of real objects, our method can closely match input appearances. More results are in Sec.~\ref{supp:more_results}.

\vspace{0.1cm}
\noindent\textbf{Single-image inputs.}
Our method can even be applied to single image inputs, as shown in Fig.~\ref{fig:teaser} and \ref{fig:results_hq} with our own \textit{Photos-to-Manual} dataset. Our framework allows material transfer for unseen portions of coarsely aligned CAD models from the photo, as shown in the novel view rendering. This makes the framework more practical than recent works \cite{nimier2021material} that require perfectly aligned geometry and multiview images to optimize for \textit{observed} geometry. In the second row of Fig.~\ref{fig:results_hq}, both \textit{material} and \textit{orientation} of the carpets with grid patterns are successfully estimated. Lastly, the estimated materials are high-resolution and photorealistic as shown in the zoom-in views and relighting results.

\vspace{0.1cm}
\noindent\textbf{Automatic 3D reconstruction and masks.}
Our method can be fully automatic by using off-the-shelf single image 3D reconstruction and panoptic (or instance) prediction for initialization. 
We illustrate this in Fig.~\ref{fig:results_total3d} with the \textit{SUN-RGBD-to-Total3D} dataset. This shows that our method is robust even when both masks and meshes are imperfect and not aligned well, which also cannot be achieved in recent works \cite{nimier2021material} that need high-quality aligned geometry.

\vspace{0.1cm}
\noindent\textbf{Variations from optimized material.}
Another advantage of procedural graph material representation is that it allows further edits from the current parameters. We can adjust material and UV parameters starting from the current estimation and generate edited results as shown in Fig.~\ref{fig:results_material_variations}. 
Note that the image becomes brighter by perturbing graph parameters under the same lighting which explains materials can also change the brightness of an image. This demonstrates the benefit of globally consistent lighting optimization with material refinement stages, which ensures materials for each part are consistent under global lighting representation.

\subsection{Baseline Comparisons}
\label{sec:baseline}

\noindent \textbf{Material classifier.} The most relevant work to ours is PhotoShape \cite{photoshape2018}, which learns a material classifier from a dataset of shapes with material assignments. The input to the network is an image with an aligned material part mask. Although PhotoShape does not consider lighting or complex indoor scenes, we compare by mimicking their approach in our setting. We borrow the material classification model from~\cite{photoshape2018} and re-train it in a whole scene setting, with classification of material parts over 886 materials and material category classification over 9 super-classes. For each input image associated with a material part mask, we predict one of 886 material labels. 
Implementation details can be found in Sec.~\ref{supp:matcls}.

\vspace{0.1cm}
\noindent \textbf{Median of per-pixel material predictions.} For this baseline, we can construct a homogeneous material from per-pixel predictions of the inverse rendering network~\cite{li2020inverse} by computing median values of per-pixel albedo and roughness under selected view for each material in the masked region and setting the normal to flat.

\vspace{0.1cm}
\noindent \textbf{Median of pixel values.} We follow IM2CAD~\cite{izadinia2017im2cad} to assign a homogeneous albedo as the median values of the 3 color channels independently within the masked region in the selected view for each material, set a fixed roughness value at $0.7$ and use a flat normal.

\begin{table}[!bt]
\begin{center}
\small
\begin{tabular}{l|cccc}
           & Classifier & InvRender Med. & Pixel Med.  \\ \hline
Ours preferred over  &  68.19\%  & 65.06\%   &  69.70\%        
\end{tabular}
\end{center}
\vspace{-0.5cm}
\caption{User study asking which method produces results more similar to the reference with ScanNet-to-OpenRooms dataset. }
\vspace{-0.3cm}
\label{tab:baseline_user}
\end{table}

\begin{table}[!bt]
\begin{center}
\small
\begin{tabular}{c|cccc}
     & Classifier & InvRend. Med. & Pixel Med. & Ours \\ \hline
RMSE &  0.452 & 0.349 & 0.337 & \textbf{0.259} \\
SSIM & 0.401  & 0.479 & \textbf{0.497} & 0.493   \\
LPIPS & 0.546  & 0.510 & 0.501 & \textbf{0.489}
\end{tabular}
\end{center}
\vspace{-0.4cm}
\caption{Similarity evaluation between baselines rendering results and reference photo with ScanNet-to-OpenRooms dataset. 
}
\vspace{-0.2cm}
\label{tab:baseline}
\end{table}

\begin{figure}
\begin{center}
\includegraphics[width=\linewidth]{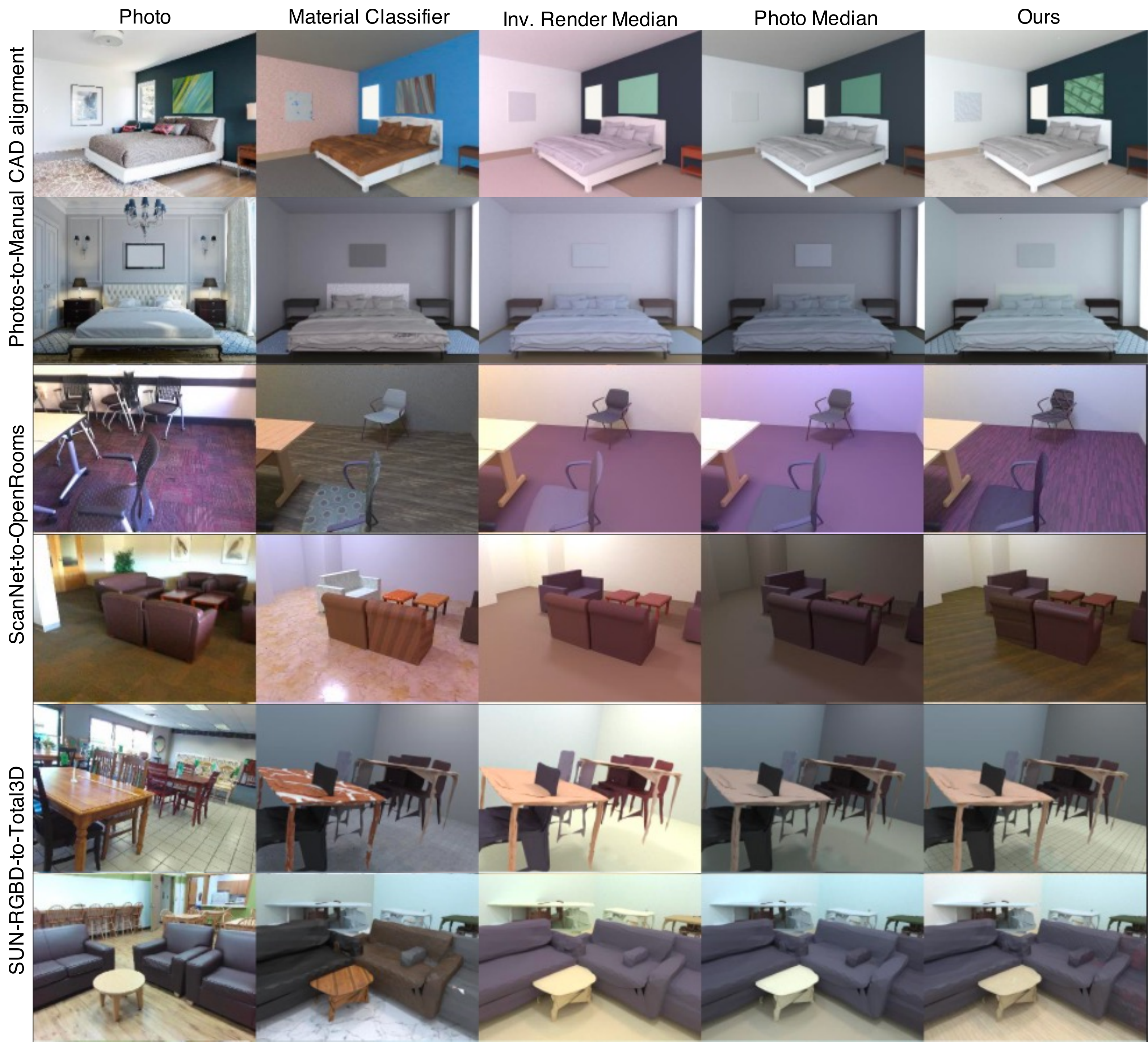}
\end{center}
\vspace{-0.5cm}
   \caption{Qualitative comparison\textsuperscript{\ref{note:comp}} with baselines on various datasets, where our method generates high-quality materials with spatially-varying patterns that better match the input photograph.  }
\vspace{-0.4cm}
\label{fig:compare}
\end{figure}

\vspace{0.1cm}
\noindent \textbf{Comparisons and user study.}
The comparisons of our results with baselines are shown in Fig.~\ref{fig:compare}
with various datasets.\footnote{\label{note:comp}As none of the baselines can estimate global lighting well, we use our predicted lighting to render baseline images to ensure fair comparison.} The material classifier can only predict material from a predefined dataset, which is not guaranteed to match the actual appearance in the photo. The median of the inverse rendering method can generate appearances close to the photo, but the albedo color sometimes goes off due to the issue of albedo-lighting ambiguities. The median of photo pixels robustly computes the albedo color similar to the photo, but both the spatially-varying patterns and the roughness are not estimated. In contrast, our method can estimate accurate spatially-varying materials which is similar to the photo as well as the global lighting. 

For quantitative evaluations, Table~\ref{tab:baseline} reports similarity metrics (RMSE, SSIM, LPIPS) between photos and renderings of various methods with 70 randomly sampled scenes, consisting of 669 material parts, using ScanNet-to-OpenRooms dataset under uniformly sampled views in Table~\ref{tab:baseline}.  
We compute RMSE on the optimized region for each material, while SSIM and LPIPS are on the entire image.
Note that these similarity metrics are not designed to evaluate similarity between misaligned images or to evaluate spatial variations, so tend to favor homogeneous outputs of the median-based methods. Nevertheless, PhotoScene outperforms all baselines on these metrics, except pixel median in SSIM. Note that the homogeneous albedo from pixel median may match a photo well on an average, but without spatial variations or accurate relighting in new views. 

To evaluate methods with human perception, we provide a user study to evaluate the similarity in Table~\ref{tab:baseline_user} using the same dataset. 
We choose 20 random scenes with uniformly sampled $4$ to $12$ views and render a set of images under selected views with our result.
We ask users on Amazon Mechanical Turk to determine which set of images is more similar to the corresponding photo set. 
More details can be found in the Sec.~\ref{supp:user_study}.
About 65 to 70$\%$ users think PhotoScene generates results more similar to the inputs. Thus, our method outperforms the baselines both qualitatively and quantitatively. 

\vspace{-0.4cm}
\paragraph{More analysis.}
We further conduct an ablation study of component choice and robustness of using different ways to obtain material part masks in Sec.~\ref{supp:ablation}.

\vspace{0.1cm}
\noindent\textbf{Discussion and Limitations.}
Our algorithm assumes a part segmentation already exists in the reconstructed geometry, and our results depend on its quality and granularity. A finer part segmentation could be achieved by retrieving objects from a higher quality CAD model collection. Our graph collection is limited to the set provided by the existing implementation of MATch~\cite{Shi2020:ToG}, though more general procedural graphs could be added with some effort, possibly using automatic techniques~\cite{hu2021inverse}. Our approach cannot handle specific patterns such as paintings, which could be addressed by training a generative model for such materials \cite{guo2020materialgan}. Lastly, we rely on a neural inverse rendering initialization, where the current state-of-the-art is restricted to small resolutions, so some high-frequency information from photos might be lost. This will likely be improved by future architectures handling higher resolutions. We note a potential negative impact of spurious edits (Deepfakes) of indoor scenes, which we discuss further in Sec.~\ref{supp:impact}.

\vspace{-0.1cm}
\section{Conclusion}
\vspace{-0.2cm}

We have presented a novel approach to transfer materials and lighting to indoor scene geometries, such that their rendered appearance matches one or more input images. Unlike previous work on material transfer for objects, we must handle the complex inter-dependence of material with spatially-varying lighting that encodes distant interactions. We achieve this through an optimization that constrains the material to lie on an SVBRDF manifold represented by procedural graphs, while solving for the material parameters and globally-consistent lighting with a differentiable renderer that best approximates the image appearances. We demonstrate high-quality material transfer on several real scenes from the ScanNet, SUN-RGBD dataset and unconstrained photographs of indoor scenes. Since we estimate tileable materials that can be procedurally generated, the scenes with transferred material can be viewed from novel vantage points, or under different illumination conditions, while maintaining a high degree of photorealism. We believe our work may have significant benefits for 3D content generation in artistic editing and mixed reality applications. Further, our approach can be used to create datasets for inverse rendering, where geometry is easier to acquire but ground truth material and lighting are hard to obtain.

\small
\noindent\textbf{Acknowledgments: }
We thank NSF awards CAREER 1751365, IIS 2110409 and CHASE-CI, generous support by Adobe, as well as gifts from Qualcomm and a Google Research Award.
\normalsize

\appendix
\section{Ablation Study}
\label{supp:ablation}
\begin{figure*}[!!t]
\begin{center}
\includegraphics[width=1.0\textwidth]{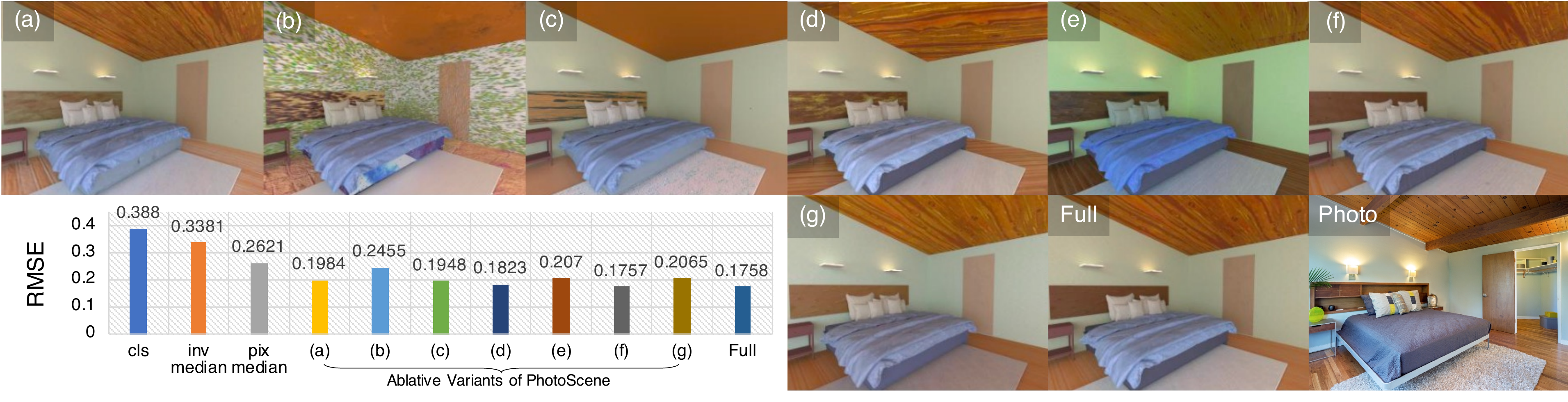}
\end{center}
\vspace{-0.4cm}
   \caption{Ablation study on our entire framework with one selected scene from Photos-to-Manual. We compare the results by removing different modules from our full framework: a) without  warping, b) with random graph selection, c) with material classifier graph selection, d) with stat loss only, e) with vgg loss only, f) without  UV transformation parameters, and g) without material reoptimization, as well as our full framework and the baselines mentioned in the main paper.}
\label{fig:ablation}
\end{figure*}

\begin{table*}[!bt]
\begin{center}
\small
\begin{tabular}{c|ccc|ccccccc|c}
      & \multicolumn{3}{c|}{Baseline Methods} & \multicolumn{7}{c|}{Ablative Variants of PhotoScene} & PhotoScene  \\ \hline
     & Classifier & InvRend. Med. & Pixel Med. & (a) & (b) & (c) & (d) & (e) & (f) & (g) & Full \\ \hline
RMSE &  0.448 & 0.381 & 0.314 & 0.250 & 0.255 & 0.251 & 0.249 & 0.326 & 0.243 & 0.272 & 0.244
\end{tabular}
\end{center}
\vspace{-0.4cm}
\caption{Similarity evaluation between rendering results and reference photo on 18 selected scenes of the ScanNet-to-OpenRooms dataset, for baseline methods, various ablations and the full version of the proposed PhotoScene approach. 
}
\vspace{-0.2cm}
\label{tab:ablation}
\end{table*}

We provide ablation study on our entire framework, by removing each component one at a time. We visualize the final results and compute RMSE on the optimized region for each material part with one scene from Photo-to-Manual dataset as shown in Fig.~\ref{fig:ablation}
and with 18 randomly selected scenes over 171 materials from ScanNet-to-OpenRooms dataset in Table~\ref{tab:ablation}. We demonstrate the results a) without warping, b) with random graph selection, c) with material classifier graph selection, d) with stat loss only, e) with VGG loss only, f) without UV transformation parameters, and g) without material reoptimization, as well as our full framework and the baselines mentioned in the main paper. 

Without warping, the mask and UV map cannot correctly fetch accurate material regions for optimization. This might lead to wrong material portions being considered due to misalignment so that the overall color and the pattern are not accurate. If we choose procedural graphs randomly from the entire collection or conditioned on a material super-class, the results do not have similar patterns as each procedural graph represents a distinct type of material (e.g. wood, homogeneous, ... etc.). With only statistics loss, the spatially-varying patterns become unconstrained and only match color statistics without considering spatial structures. The UV parameters cannot be estimated correctly and the statistics loss does not contain structure information. With only VGG loss, the results have similar spatial structures but are not guaranteed to have similar color to the reference photo without statistics loss. Without optimization of UV transformation, the orientation and scale of the textures are not guaranteed to be consistent to the reference photo. Note that even though our full method has slightly higher RMSE than (f), its qualitative superiority is not reflected in the metric since our optimization objectives are to align the pixel statistics and masked VGG features rather than per-pixel appearances. Without material re-optimization, sometimes the initial albedo colors have lighting baked-in, resulting in mismatched color under globally-consistent lighting.  It is possible to get lower RMSE values with worse UV parameters.    
In sum, our full framework generates more similar appearances to the photo by considering all the components.

\section{More Results}
\label{supp:more_results}
\begin{figure*}[!!t]
\begin{center}
\includegraphics[width=1.0\textwidth]{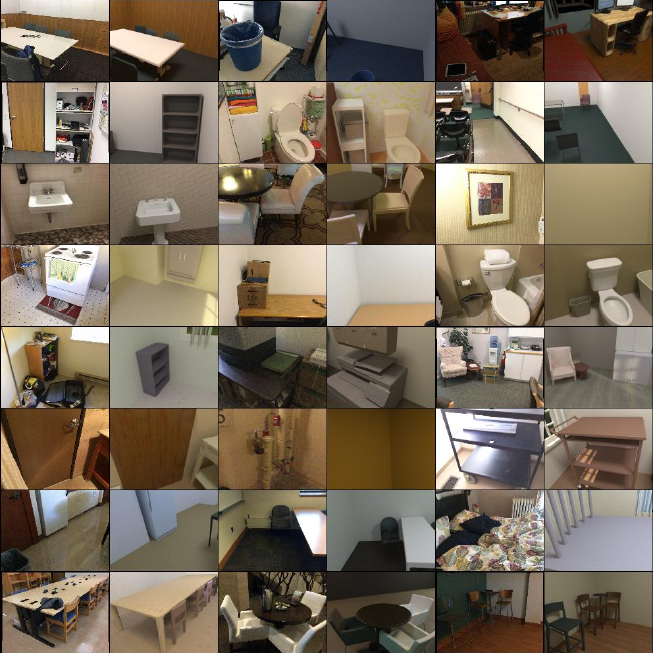}
\end{center}
  \caption{More results of material and lighting transfer with ScanNet-to-OpenRooms dataset.}
\label{fig:more1}
\end{figure*}

\begin{figure*}[!!t]
\begin{center}
\includegraphics[width=1.0\textwidth]{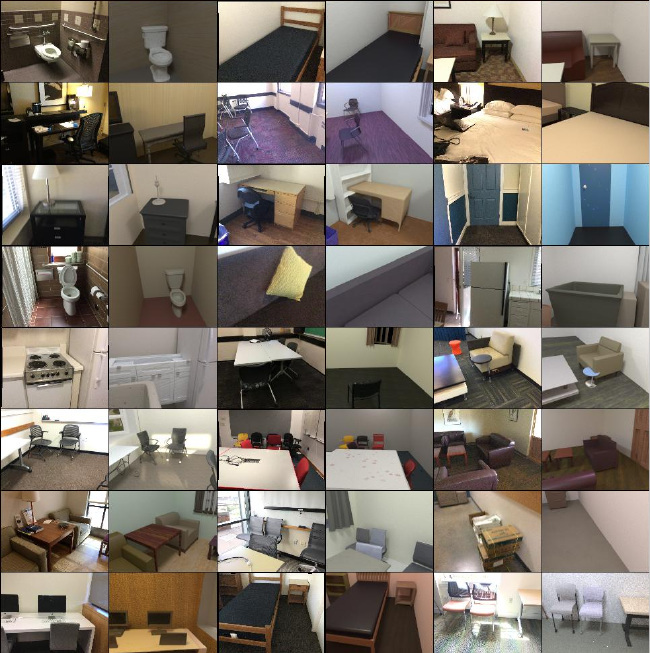}
\end{center}
  \caption{More results of material and lighting transfer with ScanNet-to-OpenRooms dataset.}
\label{fig:more2}
\end{figure*}

\begin{figure*}[!!t]
\begin{center}
\includegraphics[width=1.0\textwidth]{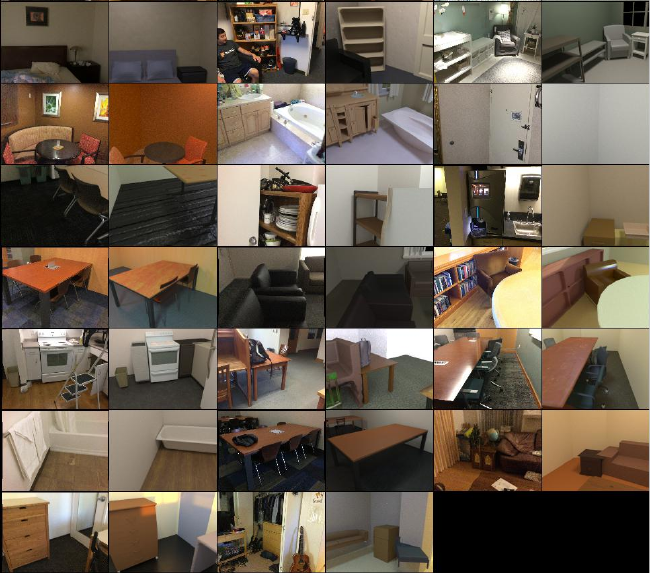}
\end{center}
  \caption{More results of material and lighting transfer with ScanNet-to-OpenRooms dataset.}
\label{fig:more3}
\end{figure*}

\begin{figure*}[!!t]
\begin{center}
\includegraphics[width=1.0\textwidth]{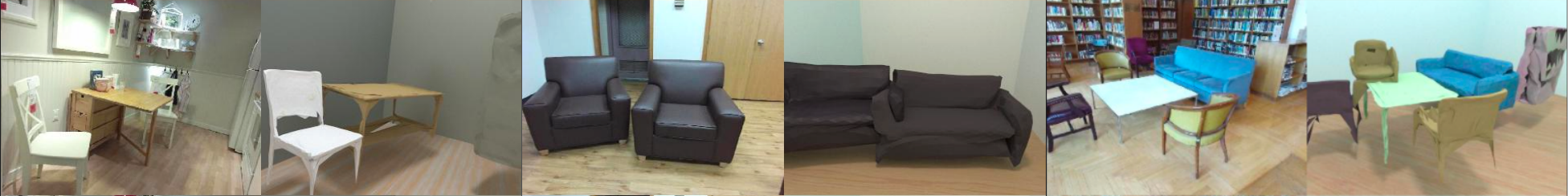}
\end{center}
  \caption{More results of material and lighting transfer with SUN-RGBD-to-Total3D dataset.}
\label{fig:more4}
\end{figure*}

We demonstrate more results on ScanNet-to-OpenRooms, Photos-to-Manual, SUN-RGBD-to-Total3D material and lighting transfer results with novel view and relighting results in 
Fig.~\ref{fig:more1},~\ref{fig:more2}, ~\ref{fig:more3}, ~\ref{fig:more4} and 
supplementary videos\footnote{Videos can be found on project page: \url{https://yuyingyeh.github.io/projects/photoscene.html}}. 

We also provide results with panoptic labels predicted by MaskFormer~\cite{cheng2021per} instead of ground truth labels for ScanNet-to-Openrooms, and compute the results with baselines with randomly selected 62 scenes and over more than 521 materials, as shown in Table~\ref{tab:mask_abl}. The RMSE errors are slightly higher when using panoptic predictions, but still lower than baseline methods with panoptic ground truths. This demonstrate that our method is robust to imperfect input mask and outperform baseline methods regardless the input masks.
\vspace{1mm}

\begin{table*}[!h]
\begin{center}
\small
\begin{tabular}{c|cccc}
     & Classifier & InvRend. Med. & Pixel Med. & Ours \\ \hline
RMSE (GT Mask) &  0.453 & 0.337 & 0.342 & 0.259 \\
RMSE (Pred. Mask) & 0.467  & 0.373 & 0.354 & 0.285 
\end{tabular}
\end{center}
\vspace{-0.4cm}
\caption{Similarity evaluation between baselines rendering results and reference photo with ScanNet-to-OpenRooms dataset using ground truth panoptic labels versus predictions from MaskFormer~\cite{cheng2021per}. 
}
\vspace{-0.2cm}
\label{tab:mask_abl}
\end{table*}

\section{Additional Details for Proposed Method}
\vspace{1mm}

\subsection{Consensus-aware View Selection}
\label{supp:consensus_select}
When a video sequence is available as input, we subsample views that are at least $30^\circ$ or $1m$ apart, then choose the optimal view among them for optimizing each material part. We choose the best view based on three criteria -- coverage, field-of-view and consensus. We expect good material transfer from those input images where a substantial number of pixels from the material part are observed. To ensure they occupy a favorable field-of-view, we weigh the number of pixels with a Gaussian, $G$, centered at the middle of the image and with variance one-fourth of the image dimensions. Finally, the goodness of a material part in a given view is also determined by the number of other views, $n_i$, where material estimates are in consensus, which is determined as the L2-norm of the mean and standard deviations of the per-pixel albedo and roughness predictions from InvRenderNet. We choose the view with the highest value of $n_i \cdot \sum (G \odot M_{photo})$ as the one to use for material transfer.

\begin{figure*}[!!t]
\begin{center}
\includegraphics[width=0.8\linewidth]{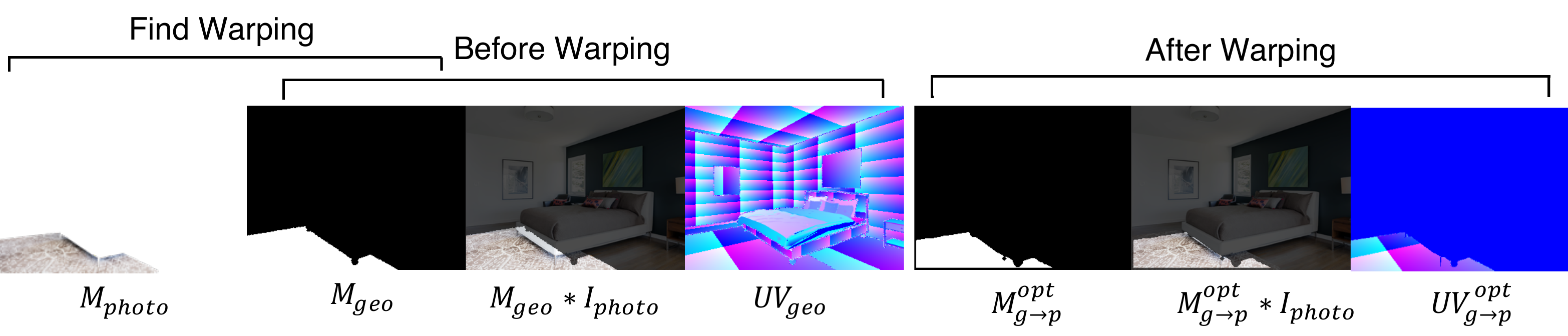}
\end{center}
\vspace{-4mm}
   \caption{Example of part segmentation matching and UV warping between geometry and input image.}
\label{fig:partMatchAndUvWarp}
\end{figure*}

\subsection{Material Part Mask and Mapping}
\label{supp:mask}
We regard material part segmentation as non-trivial, since material parts are ambiguous, e.g. table legs can be treated as separated parts or same part as the entire table. We found that the instance-based segmentation from MaskFormer already provides robust candidates which can later be refined by the mapping and alignment with geometry mask $M_{geo}$. Again, we can always provide better segmentation from manually labeling or existing dataset. 

When MaskFormer does not detect a valid mask or 3D shapes have too small parts or highly different geometries from the image, we cannot find a large enough mask. We determine these failure situations by setting a threshold on the number of valid pixels inside a mask which can be used for optimization, and simply compute median values on the valid pixels, or on geometry mask $M_{geo}$ if no valid pixels at all. 
To be specific, we first compute a per-pixel weight map $W_{aln}$ by the dot product between aligned normal from InvRenderNet $N^{inv}$ and normal from geometry $N_{geo}$ and then define the valid pixels by computing the number of pixels with the above dot product larger than $0.95$ as $J$ and only run our optimization if $J\geq500$, otherwise, we compute median for small masks where $J<500$. If there is no mask candidate being matched by IoU, we simply use $M_{geo}$ to compute median.

The weight map $W_{aln}$ is also multiplied with $M_{aln}$ to obtain a weighted mask when computing mask-based losses during optimization.

\subsection{Alignment and Warping}
\label{supp:alignment}
Let $M_{geo}$ be the 2D material part mask rendered from geometry under corresponding views and $M_{photo}$ be the mask for the reference photo. We first decompose $M_{geo}$ and $M_{photo}$ into sub-masks $\{ M_{geo}^i \}$ and $\{ M_{photo}^j \}$ which represents a single instance (if there are multiple instances), and search for matching instance pairs by the highest mIoU values on \textit{soft} instance submasks. If semantic labels for both photo and geometry are available, we can use it to reduce the sub-masks by selecting corresponding semantics. Here \textit{soft} means we apply a Gaussian filter on the instance submask with mean set as the center of mask bounding box and standard deviation set as half of width and height of bounding box, respectively.

After finding the matching pairs of part instances, we need to find the warping relationship between $M_{geo}$ and $M_{photo}$ so that we can warp $UV_{geo}$ to $\widehat{UV}_{geo} \approx UV_{photo}$, which is used to sample material parameters from $UV$ space to image space in our material part-based differentiable rendering module. We formulate the warping as scaling and translation from bounding box $B_{geo}$ of $M_{geo}$ to bounding box $B_{photo}$ of $M_{photo}$ to avoid unnecessary rotations. Let $c_{g}$ and $l_{g}$ be the center and size of $B_{geo}$ and $c_{p}$ and $l_{p}$ be the center and size of $B_{photo}$. 
While $c_{g}$, $c_{p}$ and $l_{g}$, $l_{p}$ can be computed by minimum and maximum pixel locations in $x$ and $y$ directions of $M_{geo}$ and $M_{photo}$, we can further find optimal $c_{g}^*$ and $l_{g}^*$ by optimizing intersection-over-union:
\begin{equation}
    \max_{c_{g}, l_{g} } \frac{M_{photo}\cap \widehat{M}_{geo}}{M_{photo}\cup \widehat{M}_{geo}},
\end{equation}
to ensure higher percentage of overlap between $\widehat{M}_{geo}$ (the warped $M_{geo}$) and $M_{photo}$.

We warp the $UV$ map $\widehat{UV}_{geo}$ by 
\begin{equation}
\label{eq:ptopt2}
    x_s^* = (x_t - c_{p})/l_{p}*l_{g}^* + c_{g}^*,
\end{equation}
\begin{equation}
    \widehat{UV}_{geo}(x_t) = UV_{geo}(x_s^*) \approx UV_{photo}(x_t).
\end{equation}
Finally, we derive the warped material part mask $M_{g\rightarrow p}^{opt}$ and UV map $UV_{g\rightarrow p}^{opt}$ for optimization by overlapping regions after warping:
\begin{equation}
    M_{g\rightarrow p}^{opt} = \widehat{M}_{geo}*M_{photo},\quad \widehat{M}_{geo}(x_t)=M_{geo}(x_s^*),
\end{equation}
\begin{equation}
    UV_{g\rightarrow p}^{opt} = \widehat{UV}_{geo}*M_{g\rightarrow p}^{opt}.
\end{equation}
Please see Fig.~\ref{fig:partMatchAndUvWarp} for an illustration. In the material optimization stage, $M_{aln}$ refers to $M_{g\rightarrow p}^{opt}$.

With the improved view-consistent representation of light sources, we re-optimize the materials to achieve more accurate appearance in the material reoptimization stage. However, the per-pixel lighting bakes-in the geometry in certain views, which necessitates all inputs to be aligned with the geometry. So, we warp the reference photo $I_{photo}$ and the material part mask $M_{photo}$ to match the geometry. We again define bounding box parameters $l_{p}$ and $c_{p}$ to compute the warped $\widehat{M}_{photo}$ from $M_{photo}$ to match $M_{geo}$:
\begin{eqnarray}
    &x_s^* = (x_t - c_{g})/l_{g}*l_{p} + c_{p}, \\
    &\widehat{M}_{photo}(x_t) = M_{photo}(x_s^*) \approx M_{geo}(x_t), \\
    &M_{p\rightarrow g}^{opt} = \widehat{M}_{photo}*M_{geo} ,\quad I_{p\rightarrow g}^{opt} = \widehat{I}_{photo}*M_{p\rightarrow g}^{opt}.
\end{eqnarray}
Therefore, $M_{aln}$ refers to $M_{p\rightarrow g}^{opt}$ in the material reoptimization stage.

\section{Additional Details for Experiments}
\subsection{Material Classifier Implementation}
\label{supp:matcls}
The material classification model is based on ResNet-18~\cite{he2016deep} backbone. We represent 2D convolution by \textit{Conv2D}(C, K, S, P) where C is the output channels, K is the kernel size, S is stride and P is padding. Other operations in the model include \textit{BN} for 2D batch normalization, ReLU, and Maxpool(K, S) for 2D max-pooling of kernel size K and stride S. The model takes the concatenation of the image and a binary mask of size 240$\times$320$\times$4 as input, followed by \textit{Conv2D}(64, 7, 2, 3), BN, ReLU, Maxpool(3, 2), and modules \textit{conv2.x}, \textit{conv3.x}, \textit{conv4.x}, \textit{conv5.x} from ResNet-18, and 2D average-pooling, resulting by a feature vector of dimension 512. With the feature vector as input, a fully-connected (FC) layer classifies over 886 bins of materials and another FC layer classifies over 9 super-classes. A standard cross-entropy loss is used for the classification heads.

\subsection{User Study Details}
\label{supp:user_study}
There are 60 random AMT users; each is asked to make a different binary comparison for each of 20 scenes \textit{without} pre-training on our task. In each comparison, we ask users to choose the better set of multi-view renderings of transfer results between ours and one randomly sampled baseline from {\small$\{$cls, inv med, pix med$\}$}. Two options are randomly placed while the input photo is in the middle. Each comparison is evaluated by 20 different users.

\section{Potential Negative Impacts}
\label{supp:impact}
Our approach can synthesize high-quality digital counterparts of real scenes which may be rendered to create photorealistic images. Using a physically-based material prior also allows the ability to edit properties of these images by generating plausible new materials for specific regions or objects, which may be used for potentially harmful purposes. An avenue to overcome this negative impact might be further research in digital watermarks such as \cite{stegastamp} for materials generated through our material priors, embedded in a manner that allows them to persist in an identifiable way through the rendering process.

{\small
\bibliographystyle{ieee_fullname}
\bibliography{main}
}

\end{document}